\pdfoutput=1

\documentclass[11pt]{article}

\usepackage[final]{acl}

\usepackage{times}
\usepackage{latexsym}

\usepackage[T1]{fontenc}

\usepackage[utf8]{inputenc}

\usepackage{microtype}

\usepackage{inconsolata}

\usepackage{graphicx}

\usepackage{tikz}
\usepackage{algorithm}
\usepackage{enumitem}
\usepackage{subfig}
\usepackage{amsmath}
\usepackage{makecell}

\usepackage{array}
\usepackage{needspace}

\usepackage{algpseudocode}
\usepackage{array}

\usetikzlibrary{arrows.meta, positioning, shapes.multipart, fit, backgrounds}
\usepackage[edges]{forest}
\usepackage{geometry}
\usepackage{multirow} 

\usepackage{titlesec}
\usepackage{multirow}
\usepackage{tabularx}
\usepackage{ragged2e}
\usepackage{amsfonts}

\usepackage{booktabs}

\title{Dual Hierarchical Dialogue Policy Learning for Legal Inquisitive Conversational Agents}

\author{Xubo Lin \\
  Georgetown University\\
  \texttt{xl524@georgetown.edu} \\\And
  Zezhi Deng \\
  Georgetown University\\
  \texttt{zd127@georgetown.edu} \\\And
  Shihao Wang \\
  Georgetown University\\
  \texttt{sw1379@georgetown.edu} \\\AND
  Grace Hui Yang \\
  Georgetown University\\
  \texttt{Grace.yang@georgetown.edu} \\\And
  Yang Deng \\
  Singapore Management University\\
  \texttt{ydeng@smu.edu.sg} \\}

\begin{document}
\maketitle
\begin{abstract}
Most existing dialogue systems are user-driven, primarily designed to fulfill user requests. However, in many critical real-world scenarios, a conversational agent must proactively extract information to achieve its own objectives rather than merely respond. To address this gap, we introduce \emph{Inquisitive Conversational Agents (ICAs)} and develop an ICA specifically tailored to U.S. Supreme Court oral arguments. We propose a Dual Hierarchical Reinforcement Learning framework featuring two cooperating RL agents, each with its own policy, to coordinate strategic dialogue management and fine-grained utterance generation. By learning when and how to ask probing questions, the agent emulates judicial questioning patterns and systematically uncovers crucial information to fulfill its legal objectives. Evaluations on a U.S. Supreme Court dataset show that our method outperforms various baselines across multiple metrics. It represents an important first step toward broader high-stakes, domain-specific applications.\footnote{\href{https://github.com/infosenselab/Dual-Hierarchical-Dialogue-Policy-Learning-for-Legal-Inquisitive-Conversational-Agents}{Git repository}}

\end{abstract}

\maketitle

\section{Introduction}

Conversational AI has long focused on user-driven systems suited to tasks like customer service or digital assistants. They excel when the discourse is close-ended and user–driven. However, they are not well-suited when it comes to scenarios like court justices, where they do not passively absorb information; instead, they prod, reframe, and challenge, creating a line of inquiry that tests the attorney’s narrative and hunts for latent inconsistencies. The dialogue has a moving target of questions and counters, and it is this information-seeking dynamic that we call inquisitive dialogue.

\begin{figure}[t]
    \includegraphics[width=\linewidth]{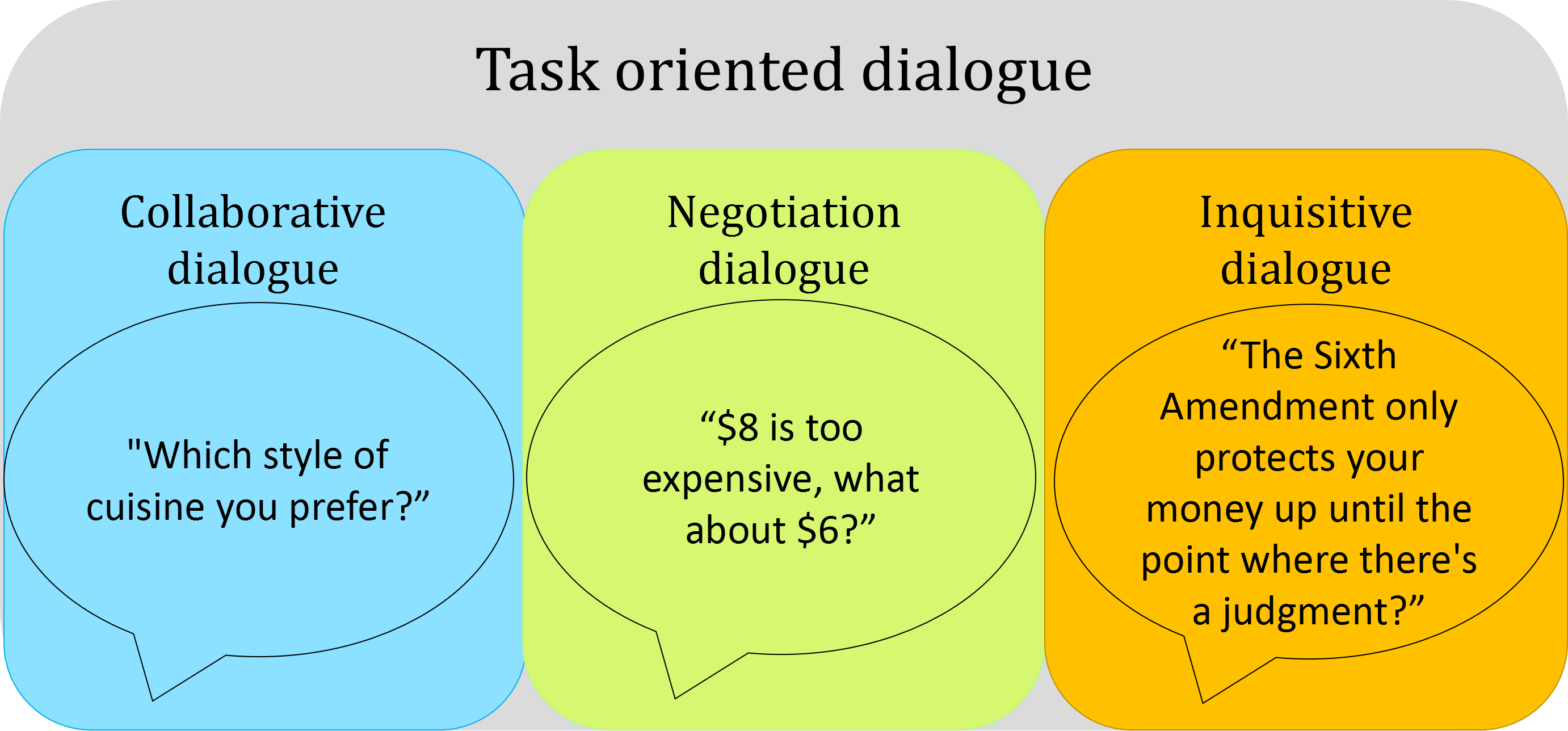}
    \caption{While this paper focuses on inquisitive dialogue in the context of U.S. Supreme Court hearings, we rethink and propose a broader categorization of task-oriented dialogue into three types: collaborative, negotiation\citep{lewis2017deal}, and inquisitive. 
    In prior works, non-collaborative types of TOD remain underexplored. 
    }
    \label{img:TOD}
\end{figure}

Much of the literature that characterizes itself as “task-oriented dialogue” in fact captures only one slice of the space: collaborative dialogue where system and user share a goal. Datasets such as MultiWOZ \citep{budzianowski2020multiwoz}, Schema-Guided Dialogue \citep{rastogi2020scalable}, Taskmaster \citep{byrne2019task}, etc., canonize that slice by framing the agent as a benevolent assistant whose sole duty is to satisfy explicit user requests. Their well-formulated slot ontologies, crowd-written templates, and short conversational arcs make them ideal for supervised learning but simultaneously ill-suited for settings where the agent, not the interlocutor, steers the agenda. Treating these resources as the entirety of task-oriented dialogue (TOD), therefore, overstates their scope and leaves the inquisitive and negotiation spectrum virtually unmapped, for example, in figure~\ref{img:TOD}, the utterance "The Sixth Amendment only protects your money up until the point where there's a judgment?" is a task oriented question but will not appear in collaborative or negotiation dialogue.

Inquisitive dialogue poses multiple challenges. First, initiative and relevance are context contingent: asking “Which soda do you prefer?” in an interview can be incisive or irrelevant depending on the preceding exchange, a nuance that traditional conversational agents can struggle to capture. Second, the interaction horizon is long. Supreme Court transcripts routinely exceed 5 000 tokens per round, stretching the capacity of mainstream encoder–decoder models that underpin many collaborative agents \citep{su2022multi, shu-etal-2019-flexibly}. Additionally, the dialogue participants do not share a common goal, and in many cases may be actively working against each other to reach their own goals. Therefore, any agent participating in inquisitive dialogue must learn long-term dialogue and questioning strategies in a non-cooperative context.

To meet these challenges, we propose a Dual Hierarchical Reinforcement-Learning (RL) framework that splits inquisitive reasoning between two tightly coupled agents. An Appraisal Agent evaluates each attorney's response in real time and converts those judgments into scalar rewards that shape the next turn, then a Hierarchical Dialogue-Policy Agent regressively generates the up to 3 hierarchies of action based on the output from the Appraisal Agent.

\section{Related Work}

\paragraph{\textbf{Proactive Conversational Agents}.}

The development of conversational agents (CAs) has been largely driven by breakthroughs in natural language processing and machine learning. Key approaches include \emph{sequence-to-sequence (Seq2Seq) modeling}~\cite{sutskever2014sequence}, \emph{pretrained language models (PLMs)}~\cite{radford2019language, liu2024llm}, \emph{retrieval-assisted text generation (RAG)}~\cite{gao2024retrieval, izacard2021leveraging}, and \emph{reinforcement learning (RL)} approaches~\cite{schulman2017proximal}. 
Among them, RL provides an optimization paradigm for dialogue strategies, particularly in \textit{task-oriented} settings~\cite{budzianowski2020multiwoz}, where reward-based learning aligns agent behavior with desired outcomes. For instance, ~\citet{li2016deep} introduced deep RL to incorporate dialogue-level rewards, while ~\citet{zhao2016endtoend} proposed an end-to-end system that learns both dialogue state tracking and strategy.

While conventional CAs typically respond to user-initiated requests, a growing line of research focuses on \emph{proactive conversational agents}~\cite{liao_proactive_2023}, which actively \emph{initiate topics}~\cite{acl19-topic}, provide \emph{context-aware recommendations}~\cite{coling20-rec}, and \emph{guide} users rather than simply reacting~\cite{emnlp23-proactive}. Proactive agents often leverage \emph{reinforcement learning}~\cite{iclr24-rl}, \emph{strategic planning}~\cite{emnlp24-strategy}, or \emph{question generation}~\cite{emnlp24-question} to address limitations of purely reactive systems, enabling richer support for tasks such as exploratory search and decision-making. ICAs take this concept even further by focusing on  \emph{steering} the conversation and \emph{gathering insights} from the user to achieve the system’s own objectives. They go beyond offering guidance or recommendations and actively \emph{probe} for information, making them especially suited to domains like legal or investigative dialogues where deeper fact-finding is critical.

\paragraph{\textbf{Legal Conversational Agents}.}

While much of the research on conversational agents has focused on open-domain or task-oriented contexts, a growing body of work explores their application in the legal domain. For instance, ~\citet{sharma2021} build a retrieval-based legal chatbot to address frequently asked legal questions. Although these systems provide valuable assistance, they predominantly adopt a reactive, FAQ-style approach, leaving vacancy for more proactive or inquisitive dialogue models—an area our work aims to advance.

\section {Problem Formulation}



\subsection{Inquisitive Conversations}
In this paper, we address the problem of \emph{inquisitive conversation}, where a conversational agent actively probes for critical information to achieve its own objectives, rather than merely responding to user queries. Specifically, we frame this challenge in the context of Supreme Court judicial dialogue. 

Inquisitive conversations
exhibit several key differences to everyday casual conversations.

\noindent
\textbf{Conversational Control:} In typical conversations, the user initiates queries and drives the topic. In judicial dialogues, the justice initiates each round of questioning and controls the direction of the discussion.

\noindent
\textbf{Purpose:} Casual dialogues often serve social or informative purposes, whereas in judicial questioning, each question aims to clarify legal uncertainty, probe for consistency, or expose logical flaws.

\noindent
\textbf{Strategy:} Justice questioning is deliberate and strategic, employing techniques such as testing hypotheticals, challenging premises, and verifying doctrinal consistency.

To model these differences in inquisitive behavior, we propose the \textbf{Inquisitive Conversational Agent (ICA)}, which mimics these questioning patterns using a dual-agent hierarchical reinforcement learning framework. 

\subsection{Dialogue Formulation}
We model the justice–attorney interaction as a Markov Decision Process (MDP), defined by the tuple $M = (S, A, R, \gamma)$, where $S$ is the dialogue state space, $A$ the action space, $R$ the reward function, and $\gamma$ the discount factor. Each dialogue round $t$ begins with a justice utterance $u_j^t$, followed by an attorney response $u_a^t$, forming an interaction pair $(u_j^t, u_a^t)$. The state $s^t \in S$ encodes the dialogue context up to round $t$.

In our formulation, the justice utterance $u_j^t$ is treated as the action $a^t$, which transitions the environment to a new state $s^{t+1}$ after observing $u_a^{t+1}$ and yields a scalar reward $r^t = R(s^t, a^t)$. 

\noindent
\textbf{Appraisal Signal:}  
In inquisitive dialogue, agents operate with their own information-seeking goals. Rather than waiting for user input to guide the exchange, they actively evaluate each response to determine whether it advances their investigative objective.
To resonate with this feature of inquisitive dialogue, we introduce an appraisal signal $p^t$ at each turn. It encodes the justice’s judgment of the attorney’s prior response (e.g., evasive, incomplete, satisfactory) under dialogue state $s^t$. 
In our dataset, the appraisal of the justice in each turn $t$ can be inferred from an utterance tuple of two rounds:
\begin{equation}
p^t = f\Bigl(u_j^{t-1}, \,
              u_a^t, \,
              u_j^t\Bigr).
\label{eq:perception-state}
\end{equation}

For instance, if the justice issues a near-identical utterance across two consecutive turns, it often indicates dissatisfaction with the attorney’s prior response. 
Accordingly, we augment the standard transition tuple to $\mathcal{D} \sim (s^t, p^t, a^t, r^t, s^{t+1})$. The Appraisal Agent treats $p^t$ as the action selected in state $s^t$, while the Dialogue Agent operates on an augmented state representation $s_{\text{aug}}^t = \text{concat}(s^t, p^t)$. 
This design enables the Dialogue Agent to condition its next action on its internal assessment of that history as well.

While many dialogue systems treat utterance generation as an open-ended natural language generation (NLG) task with a vast action space~\cite{zhao2016endtoend, sharma2017natural, Wang_2022}, domain-specific agents can often reduce complexity by operating on a finite set of \emph{dialogue acts}~\cite{peng2018deep, su-etal-2018-discriminative}. 
In the Supreme Court domain, for instance, justices frequently perform recurrent yet distinct high-level actions (e.g., asking questions, making hypotheses, or making declarations~\cite{Cichowicz2019oral}), which lend themselves to a more structured formulation. After they choose a high-level intent, such as questioning, hypothesizing, or declaring, they may refine it into a more specific subtype, like a probing or clarifying question before their actual utterance came up.

Motivated by this, we adopt a \textbf{hierarchical action space} that separates policy decisions (i.e., \emph{which dialogue act to take next}) from the lower-level surface text realization (i.e., \emph{how to verbalize that act}). Our approach discretizes justices’ interactions into a three-level taxonomy (see Table~\ref{table:SChier} in appendix) that captures both top-level acts (e.g., a “question”) and their subtypes (e.g., probing for clarity vs.\ challenging an argument). 


\begin{figure*}[t!]
    \centering
    \includegraphics[width=0.8\linewidth]{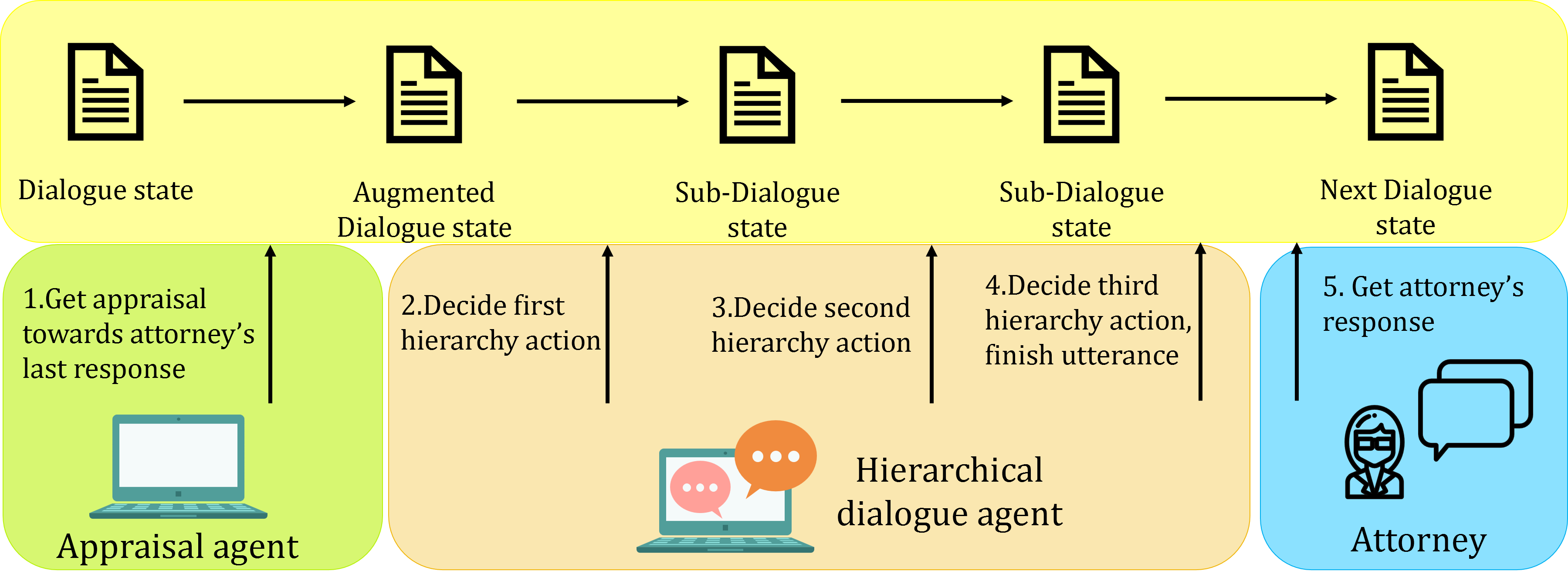}
    \caption{System Architecture of the Proposed Dual Hierarchical Inquisitive Conversational Agent.} 
    \label{fig:double_agent_flowchart}
\end{figure*}

\subsection{Reward Definition}

\label{sec:rewarddefinition}
Unlike conventional dialogue rewards that primarily assess the agent’s own utterance, our inquisitive setting focuses on how effectively the \emph{justice’s utterance} elicits information from the \emph{attorney’s subsequent response}. 
In this work, each justice's utterance \(u_{\mathrm{j}}^t\) receives a reward comprising the following components. 


\noindent\textbf{(1) Solicitation of Goal-Relevant Information.}
One objective of the agent in an inquisitive dialogue is to gather useful and relevant information that is aligned with their goal. Therefore, we introduce a goal-relevance reward to incentivize probing related to the agent's goal. To capture how effectively the justice’s utterance, \( u_{\mathrm{j}}^{t} \), compels an attorney’s response, \( u_{\mathrm{a}}^{t+1} \), to include legally significant information, we measure the attorney’s response's relevance to the case’s conclusion \( C \). Using Llama-3-8B~\cite{grattafiori2024llama3} as a semantic similarity evaluator, we compute the maximum similarity between the attorney’s response \( u_{\mathrm{a}}^{t+1} \) and each sub-conclusion \(C[i]\), with scores bounded by 5.
Formally,
\vspace{0em}
\begin{equation}
R_{\mathrm{rel}}^{t+1} (s^t, u_{\mathrm{j}}^{t} ) \;=\; \max\Bigl(\text{sim}\bigl(C[i],\, u_{\mathrm{a}}^{t+1}\bigr)\Bigr),
\end{equation}
where \(C[i]\) denotes individual sub-conclusions of the case's conclusion. 
This reward encourages justice's inquisitive utterances that steer the dialogue toward legally relevant insights.


\noindent\textbf{(2) Solicitation of Novel Information.}  
A key goal of an ICA is not only to ask questions but to drive the conversation toward uncovering information that has not yet surfaced. To capture this behavior, we introduce a \textbf{novelty reward} that measures how effectively the justice’s utterance \(u_{\mathrm{j}}^{t}\) prompts the attorney’s next response \(u_{\mathrm{a}}^{t+1}\) to contribute new and informative content beyond what has already been discussed. 
This reward encourages the agent to formulate more strategic and context-aware inquiries that elicit additional legal details or perspectives.  

Formally, we compute this reward using the \emph{Expectation-Adjusted Distinct (EAD)} metric~\cite{liu2022rethinking}, a length-normalized variant of \emph{Distinct-N}~\cite{li-etal-2016-diversity} that evaluates lexical novelty while accounting for utterance length:
\begin{equation}
    R_{\mathrm{nov}}^{t+1} (s^t, u_{\mathrm{j}}^{t}) \;=\; 
    \frac{N_{\mathrm{attorney}}^{t+1}}{
    V \left(1 - \left(\tfrac{V-1}{V}\right)^{|u_{\mathrm{a}}^{t+1}|} \right)} ,
\end{equation}
where \(N_{\mathrm{attorney}}^{t+1}\) represents the number of newly introduced tokens in \(u_{\mathrm{a}}^{t+1}\) that have not appeared in prior turns,\footnote{In the original EAD~\cite{liu2022rethinking}, \(N\) counts distinct tokens; we adapt it to track newly introduced tokens relative to the dialogue history.} \(V\) is the cumulative vocabulary size up to time \(t\), and \(|\cdot|\) denotes the token count of the utterance. 


\noindent\textbf{(3) Solicitation of Succinct Answer.}
In Supreme Court dialogues, justices often prefer brief, direct answers (e.g., “yes,” “no”) from the attorney~\cite{Cichowicz2019oral}, as these answers can swiftly confirm or deny a point and thus aid the justice’s decision-making. Additionally, succinct answers from the attorney helps the justice in keeping control of the dialogue, and conversational control is an important consideration in making an ICA. We reward this succinctness, treating it as evidence that the justice’s utterance \(u_{\mathrm{j}}^{t}\) was well-targeted:
\begin{equation}
    R_{\mathrm{clarity}}^{t+1} (s^t, u_{\mathrm{j}}^{t} )\;= \;- \,\log\!\bigl(\lvert u_{\mathrm{a}}^{t+1}\rvert\bigr),
\end{equation}
where \(\lvert u_{\mathrm{a}}^{t+1}\rvert\) is the token length of the attorney’s response. This measure complements the previous two components by explicitly encouraging \emph{clarity} in judicial exchanges.


During training, we combine the three reward components into an aggregated numerical reward via a weighted sum, which allows the agent to balance legal relevance, novelty, and clarity in its inquisitive dialogue. 

\section{Proposed Method: A Dual-Agent Framework for Legal Inquiry}
\label{sec:methodology}

Building an ICA, which actively uncovers information rather than merely answering queries, poses distinct challenges, especially in complex domains like Supreme Court hearings. To tackle this, we propose a \textit{Dual-Agent Hierarchical RL} framework, depicted in Figure~\ref{fig:double_agent_flowchart}, designed to emulate the judicial exchange process. Our approach comprises two coordinated agents, each focusing on a different aspect of the conversation.

Rather than viewing dialogue as a single flat policy, we employ a three-level hierarchical RL dialogue agent that determines \emph{when} to probe further, \emph{how} to frame questions, and \emph{if} the discussion should shift topics. By decomposing each turn into layers, ranging from broad subtopic planning to fine-grained utterance generation—the Dialogue Agent can optimize information elicitation while maintaining coherence and legal formality.

\label{sec:appraisal}
\subsection{Appraisal Agent}


We introduce a appraisal agent to \emph{evaluate} each attorney response. If the response appears evasive, contradictory, or insufficiently detailed, the appraisal agent flags the need for deeper inquiry. This mechanism mimics a justice’s tendency to monitor counsel’s answers on the fly, ensuring that the Dialogue Agent adapts its questioning in real time rather than blindly following a predefined script.

{\it Why two agents?} Separating response appraisal and dialogue control into two specialized agents enables more modular and interpretable decision-making. The Dialogue Agent focuses exclusively on planning and generating inquisitive moves, while the Appraisal Agent independently assesses whether the information obtained justifies continued exploration. 

Similar to dialogue acts, the appraisals can be discretized for a specific domain as well. We summarized nine appraisal types from Supreme Court transcripts(see Table \ref{tab:perceptionstate} in the appendix). 
These appraisals allow the justice to evaluate attorney responses, identifying flaws, seeking clarification, or prompting further inquiry, and help ensure the dialogue remains focused, responsive, and inquisitive.

In our proposed method, Appraisal Agent employs a Q-network to choose the appraisal \(p\) that maximizes its Q-value estimate:
\begin{equation}
    p(s) \;=\; \arg\max_{p}\,Q_{\text{Appraisal}}(s, p;\theta),
\end{equation}
where \(s\) is the current state embedding, and \(\theta\) denotes the Q-network parameters. The selected appraisal $p$ is then represented as a one-hot vector and merged into the Dialogue Agent’s augmented state, guiding subsequent decisions to probe further or shift to the next subtopic as needed.

In our dialogue agent, We augment the overall dialogue state $s_t$ with $p_t$ to yield $s^t_{\text{aug}} = \text{concat}(s_t, p_t)$ by treating the appraisal agent output as an internal \emph{state variable} rather than a separate action, the ICA can better track whether deeper probing is needed or if the conversation should transition to a new subtopic. 

\subsection{Dialogue Agent}
\label{sec:dialogue_agent}

To emulate Supreme Court justices, our Hierarchical Dialogue Agent first decides \emph{which} conversational act to perform (e.g., clarify, probe, or challenge), then determines \emph{how} to realize that act. We formalize these choices in a three-level action taxonomy (Table~\ref{table:SChier}). Level~1 defines high-level dialogue acts, such as \emph{Questioning}, \emph{Hypothesis Testing}, or \emph{Declaration}. Level~2 refines each act into subcategories (e.g., \emph{Clarification}, \emph{Probing}, \emph{Comparison}), while Level~3 specifies the final utterance. 

\noindent\textbf{Poincaré Embedding}
To capture the hierarchical structure of judicial dialogue acts, we represent each action in a Poincaré embedding space ~\cite{nickel2017poincare}. Poincaré embeddings are defined in a hyperbolic geometry that naturally preserves hierarchical and tree-like relationships, where parent nodes lie closer to the origin and child nodes are positioned exponentially farther away. By embedding our three-level taxonomy in this hyperbolic space, the Dialogue Agent can learn smoother transitions across levels, leverage proximity for related actions (e.g., between sibling subacts), and better generalize across hierarchically related behaviors.
The training target of it is as follows:  
\begin{equation}
\mathcal{L} = \sum_{(u,v)\in D} \log \frac{e^{-d(u,v)}}{\sum_{v' \in \mathcal{N}(u)} e^{-d(u,v')}} .
\end{equation}

Where \(d(u,v)\) denotes the hyperbolic distance between embeddings of nodes \(u\) and \(v\); \(D\) is the set of observed positive pairs (e.g., parent--child or sibling relations) derived from the dialogue act hierarchy; and \(\mathcal{N}(u)\) represents a set of negatively sampled nodes unrelated to \(u\).

\noindent\textbf{Multi-Hierarchy Action Selection.}
The three-level hierarchical action taxonomy  (Table~\ref{table:SChier}) allows our Dialogue Agent to operate at varying degrees of granularity. A single full-level action \(\{a_0, a_1, a_2\}\) may yield up to three transition tuples:
$(s,\, a_0,\, r,\, s'),$
$(s,\, a_1,\, r,\, s'),$
$(s,\, a_2,\, r,\, s'),$
The agent may terminate at any level if the chosen sub-action has no additional children.

The categories are chosen sequentially, where the highest level of action is chosen based on the augmented state, following a Level 2 action that is a subcategory of the chosen Level 1 action, and then the Level 3 action is chosen based on the Level 2 action in the same way. (e.g., choose 'question' as $a_0$, choose 'Probing question' as $a_1$, choose 'Probe the assumption' as $a_2$) Three levels of selection steps correspond to the three transition tuples above.
We use these actions to prompt LLM\citep{grattafiori2024llama3} under a unified template\ref{appendix:llm} to get a response for the Justice.

\subsection{Algorithm}
For both appraisal agent and dialogue agent, we use DDQN as the backbone, and the DDQN target for the appraisal agent is:
\begin{equation}
\label{eq:Apptarget}
Y_{\text{App}} = r \,+\, \gamma\,Q\!\Bigl(s,\, \arg\max_{p'}\,Q(s', p';\theta_{App});\theta_{App}^- \Bigr),
\end{equation}
where $\theta_{App}$ and $\theta_{App}^-$ denote the weights of the main network and the target network. The respective DDQN losses are:
\begin{equation}
\label{eq:appraisal_agent_loss}
\mathcal{L}_{\text{App}}^{\text{DDQN}} = \mathbb{E}_{\,(s, p, s')\sim \mathcal{D}} \bigl(Q(s, p;\theta_{App})\,-\,Y_{\text{App}}\bigr)^{2}, 
\end{equation}

where $\theta_{App}$ and $\theta_{App}^-$ denotes the weights of main network and target network of appraisal agent. 

For the dialogue agent, we train one Q-network that generates Q values of all possible next-level hierarchy actions sequentially conditioned on augmented states and parent actions.
The DDQN target of it is:
\begin{equation}
\label{eq:DDQNtarget}
Y_{\text{Dia}}^l = r \,+\, \gamma\,Q\!\Bigl(s,\, \arg\max_{a'}\,Q(s', a';\theta_{Dia});\theta_{Dia}^- \Bigr),
\end{equation}

where $\theta_{Dia}$ and $\theta_{Dia}^-$ denotes the weights of the main network and target network. 

We assume that the definition of dialogue actions in the dataset \(\mathcal{D}\) is complete. For any single full-level action \(\{a_0, a_1, a_2\}\), we have:
\begin{equation}
\begin{split}
\label{eq:hier}
& Q(s, a_0)=\max_{a_1}Q(s,a_1)\\
& Q(s, a_1)=\max_{a_2}Q(s,a_2)
\end{split}
\end{equation}
where \(a_1\) are all child actions of \(a_0\) and \(a_2\) are all child actions of \(a_1\). It means the Q-value of the parent action can be represented by the Q-value of the 'best' child action.
And the respective loss is
\begin{equation}
\begin{split}
\label{eq:dialogue_agent_loss}
\mathcal{L}_{\text{Dia}}^{\text{hier}} &= (Q(s, a_0)-\max_{a_1}Q(s,a_1))^2 \\
&+(Q(s, a_1)-\max_{a_2}Q(s,a_2))^2
\end{split}
\end{equation}

A well-documented challenge in offline reinforcement learning is the overestimation of Q-values for state–action pairs that are insufficiently represented in the dataset~\cite{fujimoto2019off, kumar2020conservative}. To mitigate this issue in our setting, we introduce a simple yet effective conservative regularization strategy. 
For each state \( s \), we define \( R_1(s) = \max_{a \in \mathcal{A}} Q(s, a) \), which corresponds to the maximum Q-value across all possible actions and is most likely to be overestimated. We penalize it by adding $R_1(s)$ to optimize objectives. 
However, when these high-Q actions are well represented in the dataset, applying the penalty uniformly can lead to underestimation. To address this, we introduce a compensatory term \( R_2(s) = Q(s, a) \), where \( (s, a) \in \mathcal{D} \), to restore value estimates for observed transitions.

The resulting regularization terms for the Appraisal and Dialogue Agents are defined as:
\begin{equation}
\begin{split}
\label{eq:reg}
&\mathcal{L}_{\text{App}}^{\text{Reg}} = (R_1(s)-R_2(s)) \\
&\mathcal{L}_{\text{Dia}}^{\text{Reg}} = (R_1(s_{\text{aug}})-R_2(s_{\text{aug}}))
\end{split}
\end{equation}

These terms are incorporated into the final optimization objectives for both agents as follows:
\begin{equation}
\begin{split}
\label{eq:loss}
&\mathcal{L}_{\text{App}} = \mathcal{L}_{\text{App}}^{\text{DDQN}} + \alpha\mathcal{L}_{\text{App}}^{\text{Reg}} \\
&\mathcal{L}_{\text{Dia}} = \mathcal{L}_{\text{Dia}}^{\text{DDQN}} + \beta\mathcal{L}_{\text{Dia}}^{\text{Reg}}+\lambda\mathcal{L}_{\text{Dia}}^{\text{hier}}
\end{split}
\end{equation}
Where \(\alpha\), \(\beta\) and \(\lambda\) are regularization coefficients.

When $(s, \arg\max_aQ(s,a))\in \mathcal{D}$, the regulation term is equivalent to 0. When $(s, \arg\max_aQ(s,a))\notin \mathcal{D}$, it overestimate Q-value of $(s, a)\in\mathcal{D}$ and underestimate $(s,a)$ pairs in $R_1$. This regulatory term makes the derived policy lean towards the policy that generates the dataset $\mathcal{D}$ from the potentially overestimated values. 
So by choosing appropriate $\alpha$ and $\beta$, we can reduce the variance without losing the performance.

The implementation details of algorithm can be found in Appendix \ref{alg:training}.

\section{Experiment}
\subsection{Experiment Setup}

\paragraph{\textbf{Dataset}.}
\label{sec:dataset}

We evaluate our work on the publicly available \textbf{U.S. Supreme Court Oral Argument Transcript Dataset}. In these transcripts, \emph{justices} actively probe \emph{attorneys} for information critical to deciding a case, closely reflecting the objectives of an ICA. Particularly, we use a subset of appeal court cases (spanning 1955--2023) from \url{www.Oyez.org} 
. Each transcript in this dataset contains metadata such as the case name, argument date, and speaker identifiers. The main textual content comprises a \textbf{background of the case}, an \textbf{argued question}, the complete \textbf{dialogue transcript}, and the \textbf{final conclusion}. 
Table~\ref{table:legalstats} in the appendix summarizes the distribution of cases across various legal domains. Our experiments are carried out \emph{offline}, we divide our training and evaluation data by years when the argument happened.

\paragraph{\textbf{Evaluation Metrics}.}

\label{sec:evalmetrics}

We employ two complementary evaluation strategies to assess our system’s performance. We prompt a legally pretrained, SaulLM-7B~\cite{colombo2024saullm7b}, to score each utterance generated by the agents (see Appendix \ref{appendix:llm} for the prompts). In parallel, we collect \emph{manual} ratings from human reviewers, applying the same metrics to each utterance. 

We focus on the following metrics. Both the LLM and human judges assign scores on a \emph{1–5} scale, where higher values indicate stronger performance:

\noindent
\textbf{Conformity Score (CS).}
 Measures how closely each utterance \(\{u_i\}\) reflects judicial norms (e.g., formality, legal phrasing).

\noindent\textbf{Progression Score (PS).}
  Assesses whether \(u_i\) \emph{advances} the discussion rather than stalling or digressing.

\noindent\textbf{Outcome Relevance Score (OS).}
  Evaluates each utterance’s consistency with the broader objective—such as reaching a legal conclusion or a coherent final ruling.

\noindent\textbf{Probing Effectiveness Score (PES).}
  Captures how effectively \(u_i\) \emph{prompts} new information from the interlocutor.


\paragraph{\textbf{Multi-turn Dialogue Metrics. }} We introduce two metrics to evaulate the multi-turn dialogue capabilities of the systems. We segment the original transcript from each case into topics and construct an attorney agent using \textbf{SeCom} \cite{pan2025memoryconstructionretrievalpersonalized}, and the ICA and the attorney agent engages in a simulated courtroom debate based on the question from the case. The oral argument stage has a time limit; we set the maximum conversation length as 10 rounds to represent it.

We compute a \textbf{Coverage Score} from the simulated debate, which calculates how many of the topics in the original transcript was covered by the ICA. Let $t_i$ be the original topics, $T$ be the set of topics from the simulated debate, and $t_i' \in T$ be a topic in the simulated debate. Then the Coverage Score is computed as follows:
\begin{equation}
\label{coverage_score}
    \sum_{t_i'\in T} \max_{t_i}(Sim(t_i, t_i'))
\end{equation}

We also introduce a \textbf{Marginal Relevance (MR) Score}, based on \textbf{Maximal Marginal Relevance} \cite{CarbonellGoldstein1998MMR}. The Marginal Relevance Score evaluates the ICA's ability to probe for new information while staying relevant to the topic of debate. For every round of dialogue, let $u_j$ be the justice's last utterance, and $u_{i<j}$ be the justice's previous utterances. Let $q$ be the question of the case, and $n$ be the number of dialogue rounds. Then the Marginal Relevance Score is computed as:
\begin{equation}
\label{MRS}
    \frac{1}{n}\sum_{u_j} \gamma(Sim(u_j, q)) - (1-\gamma)\max_{u_i}(Sim(u_i, u_j))
\end{equation}
We use cosine similarity for $Sim$ in both metrics. Additionally, we set $\gamma = 0.7$ to reward the justice for staying on topic, while still encouraging exploration of new topics. 

Together, these two metrics evaluate the ability of the agents to cover all the necessary topics while probing for new information in multi-turn dialogues.



\paragraph{\textbf{Baselines}.}
\label{sec:baselines}

We compare our dual-agent hierarchical RL approach against several representative conversational systems.  \textbf{Vanilla Llama3}~\cite{grattafiori2024llama3} is a straightforward \emph{prompt-only} approach, querying Llama3-8B-Instruct with no hierarchical actions or appraisals, thereby gauging the off-the-shelf capabilities of an LLM on Supreme Court discourse. \textbf{SFT Llama3}~\cite{grattafiori2024llama3} fine-tunes the same base model using our dataset, testing whether domain-specific training alone meets inquisitive dialogue demands. We also include \textbf{SaulLM-7B}~\cite{colombo2024saullm7b} to assess how specialized LLMs perform when no hierarchical or appraisal mechanisms are present. For more structured pipeline approaches, \textbf{Hudeček et al.}~\cite{hudeček2023llms} integrates domain detection, belief-state tracking, and database querying for task-oriented dialogues, while \textbf{VaRMI}~\cite{shea2023building} employs offline policy gradient and importance sampling to maintain role consistency in RL-based CAs. 
\textbf{ArCHer}~\cite{zhou2024archer} utilizes a hierarchical structure with an Actor-Critic framework for multi-turn, goal-oriented dialogues. We employ the offline variant.
Further details on hyperparameters and implementation are available in Appendix~\ref{Implementation details}.



\begin{table}[t]
\centering
\footnotesize
\begin{tabular}{|p{1.5cm}|c|c|c|c|c|}  
\hline
                  & CS   & PS   & OS   & PES   &  Overall  \\ \hline
\makecell{Vanilla\\ Llama3}           & 3.99 & 3.94 & 4.70 & 3.92 & 4.14 \\ \hline
\makecell{SFT\\ Llama3}      & 3.98 & 3.81 & 4.45 & 3.38 & 3.91 \\ \hline
\makecell{SaulLM-\\7B}         & \textbf{4.01} & 3.91 & 4.56 & 3.75 & 4.06 \\ \hline
Hudeček   & 3.99 & 3.97 & 4.77 & 3.63 & 4.09 \\ \hline
VaRMI             & 4.00 & 3.94 & 4.71 & 3.93 & 4.15 \\ \hline
ArCHer & 3.96 & 3.79 & 4.17 & 4.22 & 4.04 \\ \hline
\makecell{\bf Ours}          & \textbf{4.01} & \textbf{3.98}         & \textbf{4.89}          &  \textbf{4.47} & \textbf{4.34}  \\ \hline
\end{tabular}
\caption{Main Experimental Results 
}
\label{table:comparetobaseline}
\end{table}


\begin{figure}[t]
    \includegraphics[width=\linewidth]{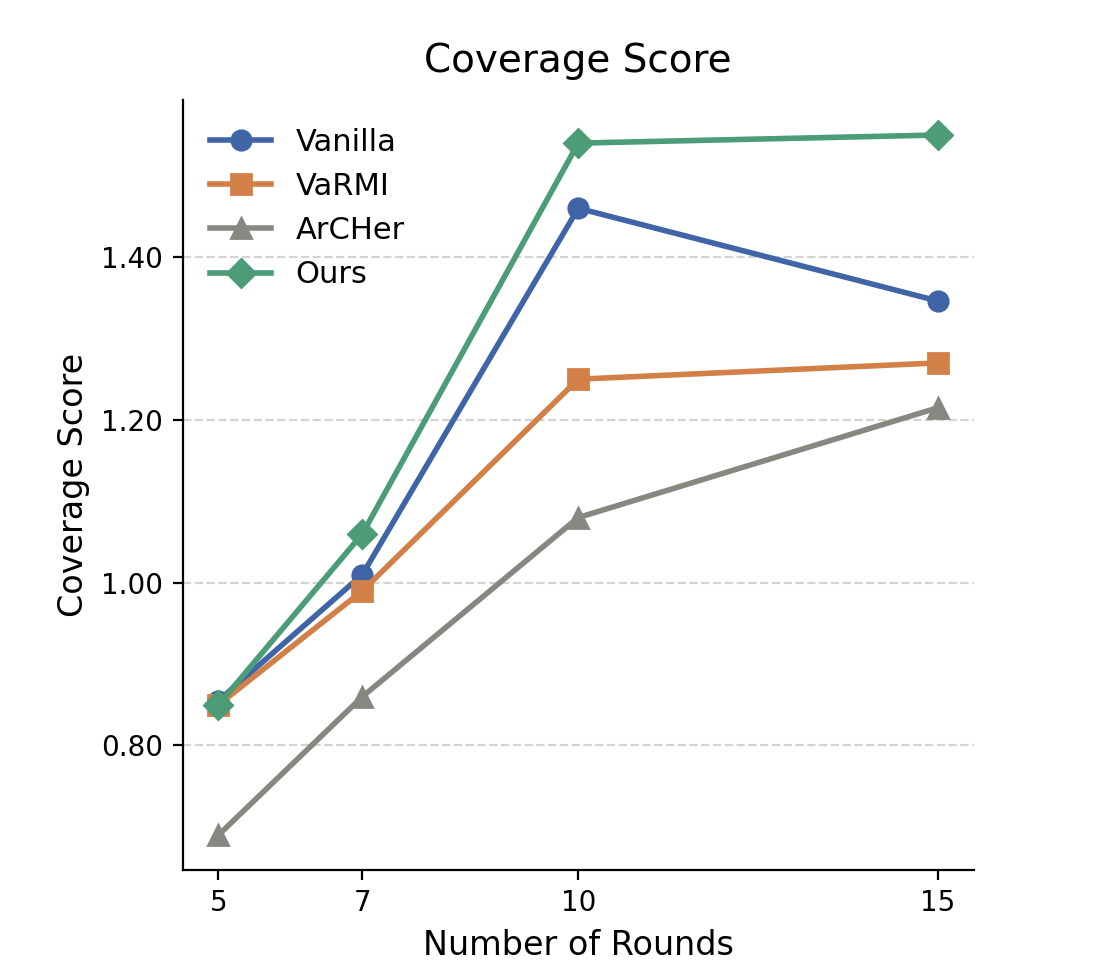}
    \caption{Coverage Score results
    }
    \label{img:Cov_score}
\end{figure}

\begin{figure}[t]
    \includegraphics[width=\linewidth]{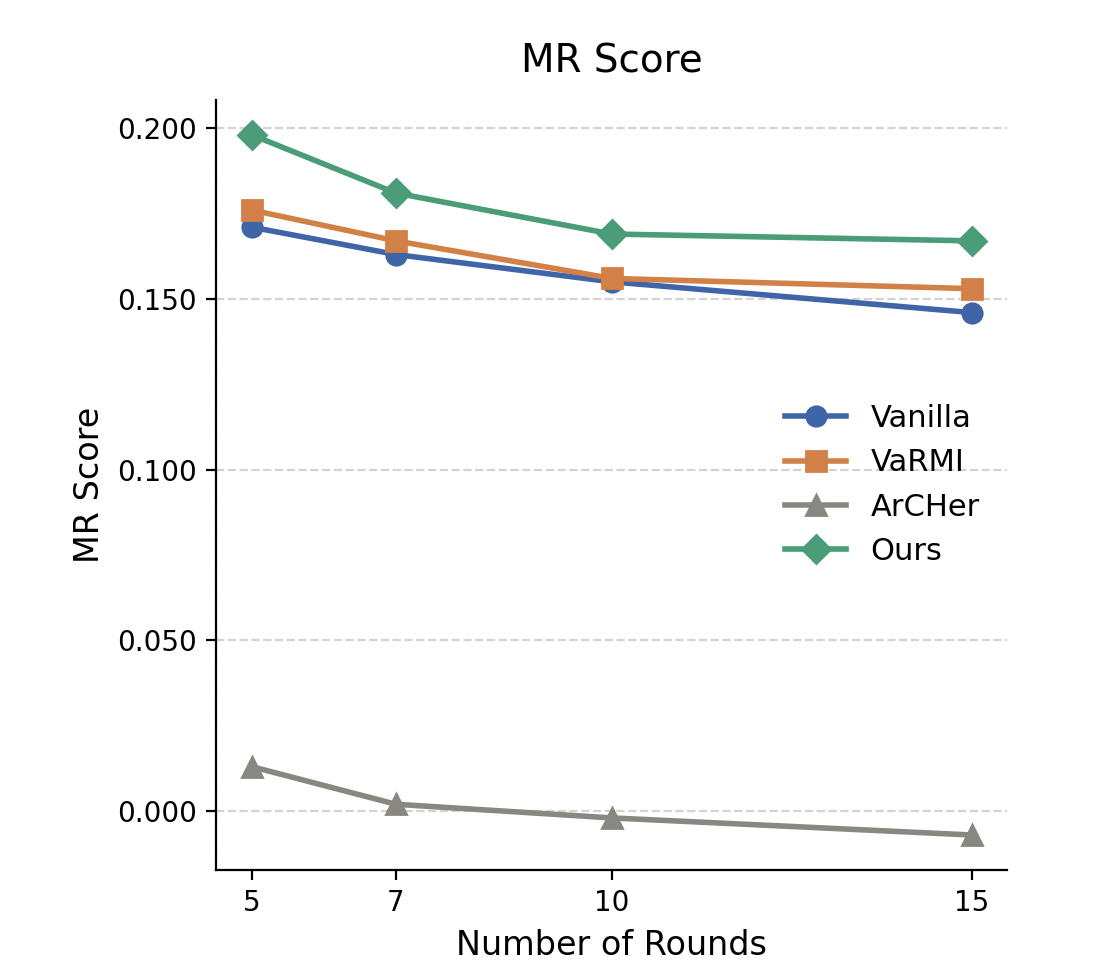}
    \caption{MR Score results
    }
    \label{img:MR_score}
\end{figure}

\begin{table}[t]
\centering
\footnotesize
\begin{tabular}{|p{1.6cm}|c|c|c|c|c|}
\hline
                          & CS            & PS           & OS            & PES            & Overall       \\ \hline
\textbf{Full Model}       & 4.01 & 3.98         & \textbf{4.89}          & \textbf{4.47} & \textbf{4.34} \\ \hline
\makecell{w/o \\Appraisal \\Agent}       & \textbf{4.03}          & \textbf{4.0} & 4.74          & 4.30          & 4.27          \\ \hline
\makecell{w/o Succinct \\Reward }      & 4.01          & 3.97         & 4.85          & 4.39 & 4.31          \\ \hline
\makecell{w/o Novelty \\Reward }       & 4.01          & 3.97         & 4.82          & 4.34          & 4.29          \\ \hline
\makecell{w/o Goal \\Relevance} & 4.00 & 3.97         & 4.83 & 4.32          & 4.28          \\ \hline
\end{tabular}
\caption{Ablation Study}
\label{table:ablation}
\end{table}

\subsection{Main Results}
In this section, we test our method and all baselines on the US Supreme Court dataset and compare their effectiveness in terms of the evaluation metrics. Detailed results are shown in Table \ref{table:comparetobaseline}. Our fine-tuning-free method achieves the best performance across all metrics, confirming that our dual agent method understands the goal of justice and its inquisitive nature well. The appraisal agent contributes to the PES metric the most, which is the metric where our method outperforms the baseline the most.

It is worth noting that although the US Supreme Court transcript is included in the training set of SaulLM-7B, it is still outperformed by generic models. The reasons for this phenomenon are two-fold: first, the model wasn’t trained for dialogue tasks; second, the task is substantially more challenging than the metrics used for SaulLM-7B.

The results of \textbf{Coverage Score} and \textbf{MR Score} are presented in Figure \ref{img:Cov_score} and Figure \ref{img:MR_score}. In Figure \ref{img:Cov_score}, our method consistently achieves the highest Coverage Score across all round settings, indicating that our dual-agent framework is more effective at expanding the discussion to cover a broader range of case-related topics. Figure \ref{img:MR_score} shows a similar pattern for MR Score: our method maintains the strongest marginal relevance throughout, suggesting that it is better able to introduce new information while remaining aligned with the central question of the case.

Due to the quality issue of the Surpeme Court dataset, the finetuning methods are not efficient on the dataset(see examples in Table \ref{low_quality}). SFT and ArCHer achieve ideal results in CS and PES, however, their results were affected by the widespread presence of low-quality data, while our approach effectively bypasses low-quality snippets.

\subsection{Ablation Study}
\label{sec:ablation}

We conducted four ablations to clarify the role of each reward component and the appraisal agent: 
(i) w/o the appraisal agent, 
(ii) w/o the succinct reward, 
(iii) w/o the novelty reward, 
and (iv) w/o the goal relevance reward. 

Table~\ref{table:ablation} shows that each omission reduces at least one key metric relative to our full model, which yields the highest overall score (4.34), confirming that all components contribute to overall effectiveness. For example, removing the novelty reward reduces OS from 4.34 to 4.29, suggesting that without encouraging fresh information, the dialogue risks becoming less directional.


Figure~\ref{fig:ablation_short} (130 epochs) and \ref{fig:ablation_long} (1600 epochs) plots the cumulative reward during offline RL. Early in training, the full model quickly surpasses ablations, reflecting the synergy of dual-agent oversight and the combination of all reward signals. 

\subsection{Human Evaluation}
We conducted a human evaluation, giving annotators the metadata of each Supreme Court case along with its dialogue context. Evaluators scored the Conformity Score (CS), Progression Score (PS), Outcome Relevance Score (OS), and Probing Effectiveness Score (PES) on a 1--5 scale (Section~\ref{sec:evalmetrics}). To ensure consistency, all methods were evaluated on the same set of case transcripts.

Table~\ref{table:human_evalutaion} presents the average ratings. Our full model achieves the highest overall score (4.53), outperforming both SaulLM-7B~\cite{colombo2024saullm7b} and all ablated versions. This underscores the importance of every component in the agent in improving performance. 




\section{Conclusion} 
\label{sec:conclusion}

In this paper, we revisit the scope of TOD and propose a three-way categorization—collaborative, negotiation, and inquisitive dialogue—to better capture the diversity of goal-driven conversation. Our study centers on the inquisitive dialogue setting, using U.S. Supreme Court oral arguments as a representative domain.

We presented a dual-agent hierarchical RL approach for inquisitive conversation, focusing on U.S. Supreme Court oral arguments as a high-stakes domain. By integrating a Hierarchical Dialogue Agent that decomposes conversation control across multiple levels with an Appraisal Agent that proactively evaluates attorney responses, our framework captures the justice’s goal-driven and probing style. We also present a regulation term that efficiently reduce the variance of our offline RL method. Empirical results on diverse Supreme Court cases show that the dual-agent design, coupled with carefully designed reward components yields more effective and context-aware dialogue strategies than multiple baselines.

While our current work centers on Supreme Court interactions, the underlying principles, such as active inquiry, structured dialogue management, and reward-driven question formulation, are broadly applicable to other high-stakes or domain-specific settings where deeper questioning is crucial. Future directions include expanding the reward model to capture even more nuanced legal strategies, and adapting the framework to other inquisitive domains such as investigative journalism or medical consultations.

\section{Limitation}
The simulated justice's responses of our agents are given by prompting LLM. Our agent's capability is heavily relies on capability of LLM.
When LLMs have a very low probability of generating the desired optimal sequence, our method cannot reach optimal performance as well. 

Although our work outperforms other baselines on the US Supreme Court dataset, the efficiency of our method on other legal domain dialogue datasets remains unclear. Our reward signals and action types are set for this dataset; for other datasets, they have to be redesigned. The policy of generating the US Supreme Court dataset is close to the optimal policy. When the dataset contains a large amount of data generated by a bad policy, our regularization term could be less efficient.

\section{Ethical statement}
This study uses publicly available transcripts and metadata from U.S. Supreme Court oral arguments, all of the original format of data can be download from \href{https://www.supremecourt.gov/}{Official website}\citep{supremecourt2025}.
The Court releases transcripts as part of its routine transparency practices.
These datasets do not reveal any identifiable information about the raters. We are not asking for any personal information during the labeler selection and labeling process. We do not include any personalized information in data processing. All of the examples used in prompting are randomly selected.

\section*{Acknowledgments}
This research was supported by U.S. National Science Foundation grant number IIS-2336768. Any opinions, findings, conclusions, or recommendations expressed in this paper are of the authors and do not necessarily reflect those of the sponsor.

\bibliography{custom}

\appendix

\section{Implementation details}
\label{Implementation details}




\paragraph{Implementation Details.}
We define the agent’s \textbf{state $S$} at each time step as the dialogue context up to the current turn. 
We transform the components into a dense vector, $S = \text{Embed}(s_c, s_h)$, using a fine-tuned Mistral-7B model~\cite{wang2024improving}. 
Our embedding model produces 4096-dimensional vectors. In both the hierarchical dialogue agent and reward model, we compress these embeddings to 32 dimensions before concatenating them with appraisals and actions. This compression uses fully connected layers with batch normalization and Leaky ReLU activations.

During training, our Dual Hierarchical Dialogue Agent relies on ground-truth appraisal and three hierarchies from the dataset, so the appraisal agent and three hierarchies of dialogue agent can be trained simultaneously while loading the same dataset.

To stabilize training, we employ polyak updates of the target networks with $\tau = 0.005$, and empirically set the discount factor $\gamma$ to 0.9. The weights for relevance, novelty, and succinctness rewards are set to 0.2, 0.7, and 0.1, respectively. 
We use exponential decay for learning rate of both agents; the learning rates for them are 1e-6 to 3e-9 and 1e-6 to 1e-8.
The model size for both appraisal agent and dialogue policy agent are both less than 2M, the whole training time takes approximately 70 hours.

Our method, ablations, SFT Llama3, VaRMI, Hudeček's method, and the vanilla baseline, use Llama-3.1-8B-Instruct as their base model. For ArCHer, the actor uses Llama-3.2-1B-Instruct as the base model due to memory constraints, and the critic uses RoBERTa, in line with the implementation presented by the authors. We run SFT Llama3 via LLaMA-Factory~\cite{zheng2024llamafactory} on 2,000 examples from the Supreme Court dataset, training for three epochs with LoRA~\cite{hu2022lora} at a per-device batch size of 4, a gradient accumulation step of 8, and a learning rate of $1\text{e}{-4}$. For VaRMI~\cite{shea2023building}, we fine-tune with a $1\text{e}{-6}$ learning rate for one epoch. In Hudeček’s method~\cite{hudeček2023llms}, the training dataset serves as the retrieval database, and cosine similarity on embedded contexts is used as the retrieval similarity function. For ArCHer, we train with a dataset of 700 dialogue trajectories, each with 6 to 7 dialogue turns, for 8 epochs, with an actor learning rate of $1\text{e}-4$ and a critic learning rate of $1\text{e}-5$.

\noindent\textbf{Inference Phase.} 
At inference, the Appraisal Agent is invoked first to generate an appraisal, which is subsequently fed into the dialogue agent. The dialogue agent first selects a top-level macro-action and then refines it through second- and third-level choices according to Table~\ref{table:SChier}, stopping if the current sub-action has no successor. Formally, each level \(l\) solves:
\begin{equation}
     a_l = \arg\max_{a_l}\,Q^{(l)}\bigl(s_{\text{aug}},\,a_0,\dots,a_{l-1},a_l\bigr).
\end{equation}
The final action vector \(\{a_0, a_1, a_2\}\) thus encodes a context-aware dialogue strategy, navigating the conversation at multiple levels of granularity
.

\subsection{Reward Model for Offline Evaluation}
\label{sec:rewardmodel}

In this work, we use an offline RL setting, and the environment is not accessible for providing real-time feedback. We thus learn a \emph{Reward Model} to approximate the environment’s true reward function from the offline training data. Particularly, we employ a feed-forward neural network (FFN) to model $R_{\phi}$ and predict a scalar reward \(\hat{r}\). The network takes data tuples of $(s, p, a, r)\sim \mathcal{D}$. 
A standard mean-squared error (MSE) loss is used here to measure the discrepancy between the predicted reward \(\hat{r}=R_{\phi}\bigl(z,\,p,\,a\bigr)\) and the ground-truth \(r_i\).

\begin{algorithm*}[t!]\small
\RaggedRight
\caption{Dual Hierarchical RL (Offline Training Mode with Regularization)}
\label{alg:training}
\begin{algorithmic}[1]
    \Statex \hspace{-\algorithmicindent}\textbf{Input:} dataset $\mathcal{D}\sim((s, p), a, r, s')$
    \Statex \hspace{-\algorithmicindent}\textbf{Output:} Policies for dual agent $\theta_{\text{App}}$ and $\theta_{\text{Dia}}$
    \State \textbf{Initialization:} Build dataset $\{(s, p, r, s')\}$ for appraisal agent, 
    \(\{\, ((s, p, a_0, \ldots, a_{l-1}),\, a_l,\, r,\, s') \mid l \in \{0,1,2\} \,\}\) for hierarchical agent, $\{(s, p, a, r)\}$ for reward model. Initialize policy and target networks \(Q_\text{App}^{\theta}\), \(Q_\text{Dia}\), and reward model \(R_{\phi}\).

    \medskip
    \Statex \textbf{Train Reward Model:}
    \For{each training iteration}
        \State Sample mini-batch $(s, p, a, r)$
        \State Update $R_{\phi}$ by minimizing
        $\mathcal{L}_{RM} = \mathbb{E}_{(s, p, a, r) \sim \mathcal{D}}\Bigl[(\hat{r}-r)^2\Bigr]$
    \EndFor \: when $R_{\phi}$ converges

    \medskip
    \Statex \textbf{Train Appraisal Agent:}
    \For{each training iteration}
        \State Sample $(s, p, r, s')$
        \State Compute DDQN target:
        $Y = r + \gamma Q_{\text{App}}(s', \arg\max_{p'} Q_{\text{App}}(s', p';\theta);\theta^-)$
        \State Compute regularization terms:
        $R_1(s) = \max_{p'} Q_{\text{App}}(s, p'), \quad R_2(s) = Q_{\text{App}}(s, p) \text{ where } (s, p) \in \mathcal{D}$
        \State Update \(Q_{\text{App}}\) by minimizing:       $\mathcal{L}_{\text{App}} = (Q_{\text{App}}(s, p) - y)^2 + \alpha (R_1(s) - R_2(s))$
    \EndFor \: when \(Q_{\text{App}}\) converges

    \medskip
    \Statex \textbf{Train Hierarchical Dialogue Agent:}
    \For{each training iteration}
        \State Sample transitions \(\{\, ((s,p,a_0,\dots,a_{l-1}), a_l, r, s') \mid l\in\{0,1,2\} \,\}\) from \(\mathcal{D}\)
        \State Compute DDQN target:
        $y^l = r + \gamma\, Q_{\text{Dia}}(s', \arg\max_{p'} Q_{\text{Dia}}(s',p';\theta); \theta^-)$
        \State Compute conservative terms:
        $R_1(s_{\text{aug}}) = \max_{p'} Q_{\text{Dia}}(s_{\text{aug}}, p'), \quad R_2(s_{\text{aug}}) = Q_{\text{Dia}}(s_{\text{aug}}, p)$
        \State Compute $Q(s, a_0), \max_{a_1}Q(s,a_1), Q(s, a_1), \max_{a_2}Q(s,a_2)$ (depends one hierarchy depth)
        \State Update $Q_{\text{Dia}}$ based on accumulate regularized loss: $\mathcal{L}=\sum_{i=1}^3\mathcal{L}^l$, where
        $\mathcal{L}^l = (Q_{\text{Dia}}(s_{\text{aug}}, p) - y^l)^2 + \beta (R_1(s_{\text{aug}}) - R_2(s_{\text{aug}}))$
        \State Offline evaluation: 
        $\hat{r} = \mathbb{E}_{(s, p, a, r) \sim \mathcal{D}} R_{\phi}(s, p^*, a_0^*, a_1^*, a_2^*)$
    \EndFor \: when \(\mathcal{L}\) converges and \(\hat{r}\) stabilizes
\end{algorithmic}
\end{algorithm*}

\begin{table*}[t]
\centering
\setlength{\tabcolsep}{2pt}
\renewcommand{\arraystretch}{0.99}
\begin{tabular}{|p{6cm}|p{9.2cm}|}
\hline
\multicolumn{2}{|c|}{\textbf{Question}} \\
\hline
\multirow{3}{*}{\textit{Clarification question}} 
 & Clarify important aspect of the case \\\cline{2-2}
 & Clarify legal arguments or issues \\ \cline{2-2}
 & Clarify definition of concept \\
\hline
\multirow{2}{*}{\textit{Probing question}}
 & Probe the consistency between the attorney's arguments and established legal principles or precedents \\ \cline{2-2}
 & Probe the assumption underlying the attorney's arguments \\
\hline
\multirow{3}{*}{\textit{Leading question}}
 & Ask for the attorney's position \\ \cline{2-2}
 & Lead the attorney toward a particular conclusion \\\cline{2-2}
 & Lead the attorney to certain aspects \\
\hline
\multicolumn{2}{|c|}{\textbf{Make hypothesis}} \\
\hline
 \multirow{2}{*}{\parbox{2.8cm}{\textit{Present hypothesis}}}
 & Present hypothetical situations to test legal limits \\ \cline{2-2}
 & Present hypothetical situations to test legal issues in the case \\
\hline
\multirow{2}{*}{\textit{Compare hypothesis}}
 & Compare to hypothetical situations to assess legal principles \\ \cline{2-2}
 & Highlight key differences from hypothetical situations \\
\hline
\textit{Conclude hypothesis}
 & Explore different types of consequences \\
\hline
\multicolumn{2}{|c|}{\textbf{Declaration}} \\
\hline
\multirow{2}{*}{\parbox{2.8cm}{\textit{Confirmation}}}
 & Acknowledge the attorney's arguments \\ \cline{2-2}
 & Prompt for information that would support the attorney's arguments \\
\hline 
\textit{Rejection}
 & Oppose the attorney's arguments \\ \cline{2-2}
 & Provide counterexample to challenge the attorney's arguments \\ 
\hline
 \textit{Declaration (non-questions) for more details}
  & Lead attorneys by examples (non-questions) for detailed explanation of a concept \\
 \hline
\textit{Declaration with Time Pressure}
 & Pressure a rash response from the attorney \\
\hline

\end{tabular}
\caption{Proposed Hierarchy of Dialogue Acts}
\label{table:SChier}
\end{table*}

\section{Three-Level Taxonomy of Justice's Action}
The hierarchies of dialogue actions are listed in Table \ref{table:SChier}. Actions in primary hierarchies are bolded. Actions in second and third hierarchies are listed in the left and right columns of the table, respectively.

\begin{table}[ht]
\centering
\clearpage
\begin{tabular}{@{}p{0.25\columnwidth} p{0.68\columnwidth}@{}}
\toprule Appraisals
 & \textbf{Explanation}\\
\midrule
Sense ambiguity & The justice finds the attorney's arguments ambiguous\\
\addlinespace[4pt]
Find deviates    & The justice believes the conversation has strayed into irrelevant territory or unproductive arguments\\
\addlinespace[4pt]
Find redundancy  & The justice finds the attorney's arguments repetitive or unproductive\\
\addlinespace[4pt]
Spot weakness      & The justice spots a weakness in the attorney's arguments\\
\addlinespace[4pt]
Identify flaws      & The justice identifies logical flaws in the attorney's arguments\\
\addlinespace[4pt]
Identify chances    & The justice identifies chances to influence the attorney \\
\addlinespace[4pt]
Keep challenging & The justice wants to challenge the attorney from another aspect\\
\addlinespace[4pt]
Dive deeper   & The Justice wants to dive deeper into the dialogue\\
\addlinespace[4pt]
Otherwise & All other kinds of the justice intents\\
\bottomrule
\end{tabular}
\caption{Appraisal Actions}
\label{tab:perceptionstate}
\end{table}

\begin{table}[!ht]
\centering
\setlength{\tabcolsep}{0.5pt}
\renewcommand{\arraystretch}{0.99}
\footnotesize
\begin{tabular}{@{}lccrr@{}}
\toprule
\textbf{Domain}
  & \textbf{Turns/case}
  & \textbf{Words/utter}
  & \textbf{Words/case}
  & \textbf{\#Cases} \\
\midrule
Regulatory   & 192.6 & 47.7 & 9183.2  & 589 \\
Civil Rights & 206.0 & 44.0 & 9074.5  & 337 \\
Criminal     & 200.6 & 45.0 & 9035.7  & 418 \\
IP           & 173.4 & 52.5 & 9101.2  & 19  \\
Commerce     & 199.3 & 46.4 & 9252.2  & 107 \\
Labor        & 262.3 & 54.9 & 14407.3 & 101 \\
Immigration  & 176.4 & 55.3 & 9762.3  & 16  \\
Environment  & 262.3 & 54.9 & 14407.3 & 3   \\
Others       & 200.4 & 47.6 & 9541.2  & 18  \\
\textbf{Total} & 198.6 & 45.9 & 9121.5 & 1608 \\
\bottomrule
\end{tabular}
\caption{Supreme Court Dataset Statistics}
\label{table:legalstats}
\end{table}

\clearpage
\section{Details Regards Measurements}
We have qualified students for making human evaluations; we collect our human evaluation results by distributing Google Forms. The estimated payment is 20\$ per hour.

\begin{table}[ht!]
    \centering \small
    \begin{tabular}{p{1.5cm} p{5.5cm}}
    \centering \textbf{Component} & \centering \textbf{Prompt} \tabularnewline
    \hline
        Instruction & You are a duteous, respectful, and honest AI justice assistant. You are given a clip of dialogue that happens on the US supreme court appeal case ending with the utterance of the justice, your job is provide analysis and score the last utterance of the justice in the dialogue from different aspect with the max score of 5. In each task, you are given the background, argued question of the case, conclusion of the case and the justice utterance with related dialogue context. Provide a snippet of analysis to analyze the role of the last sentence before giving out the score.
\\
\hline
        Metric \newline Explanation & The metric is: \newline \{Metric\}: \{Explanation of metric\} \newline \newline The scoring format should be: \newline \{Metric\}: \{Comment about \{metric\}\} ?/5\\ \hline
        Score \newline Definition & The definition of scores are: \newline Score 1/5 (\{Descriptor\}): \{Explanation\}\newline Score 2/5 (\{Descriptor\}): \{Explanation\}\newline Score 3/5 (\{Descriptor\}): \{Explanation\}\newline Score 4/5 (\{Descriptor\}): \{Explanation\}\newline Score 5/5 (\{Descriptor\}): \{Explanation\} \\
\hline
    \end{tabular}
    \caption{Prompt Structure for LLM Evaluation}
    \label{appendix:llm}
    \label{tab:llmeval}
\end{table}


\begin{figure}[ht!]
    \centering
    \includegraphics[width=0.9\linewidth]{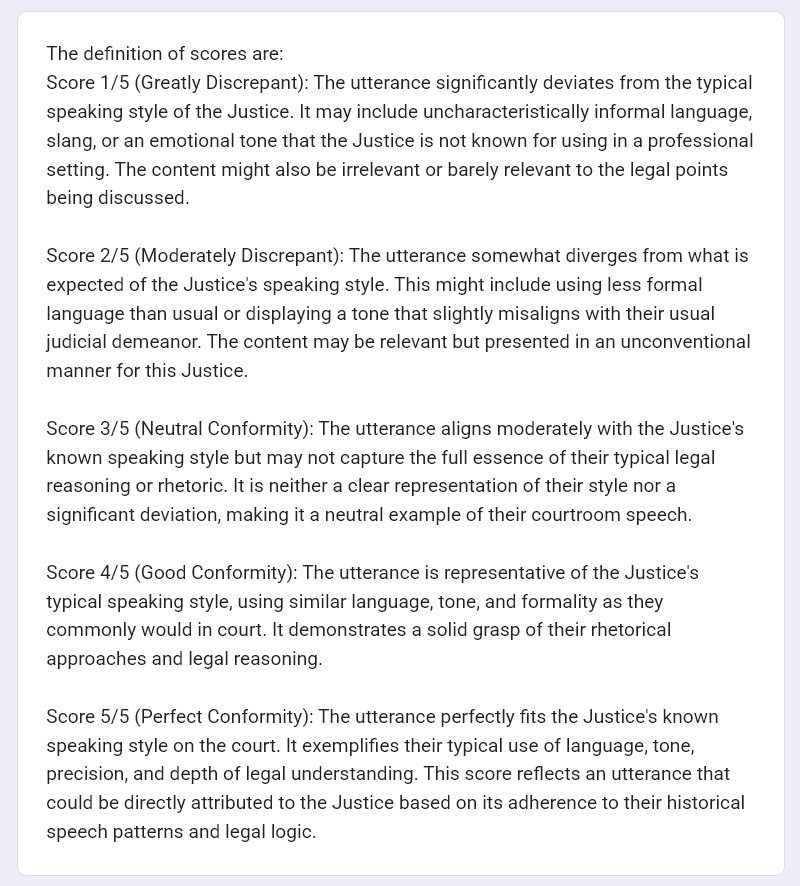}
    \vspace{1em} 
    \includegraphics[width=0.9\linewidth]{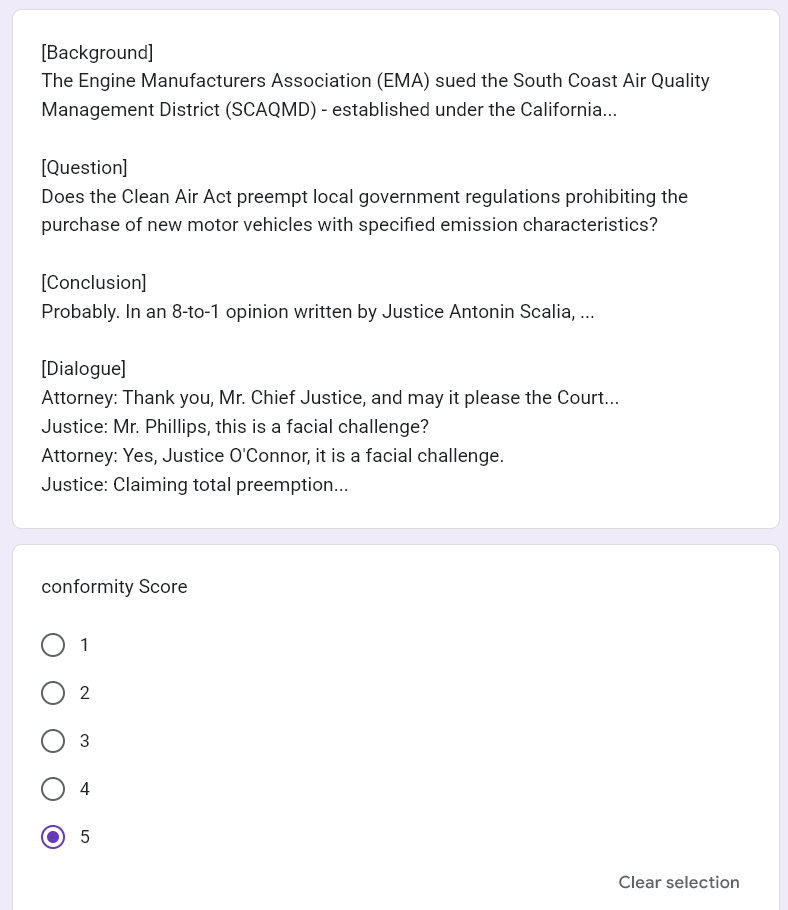}
    \\
    \caption{A template of google form for manual labeling, text has been streamlined for typographical purposes.}
    \label{fig:two-vertical}
\end{figure}

\newpage

\clearpage
\begin{figure}[ht]
    \includegraphics[width=\linewidth]{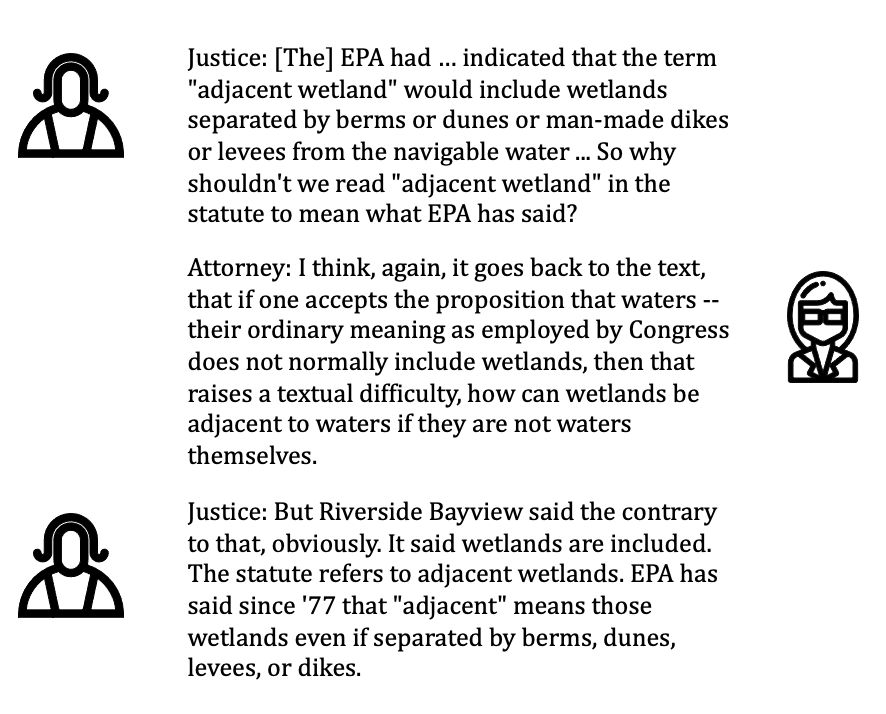}
    \caption{Justice uses a counterexample to challenge the attorney's position and the arguments presented previously
    }
    \label{img:perception2}
\end{figure}

\begin{figure}[ht]
    \includegraphics[width=\linewidth]{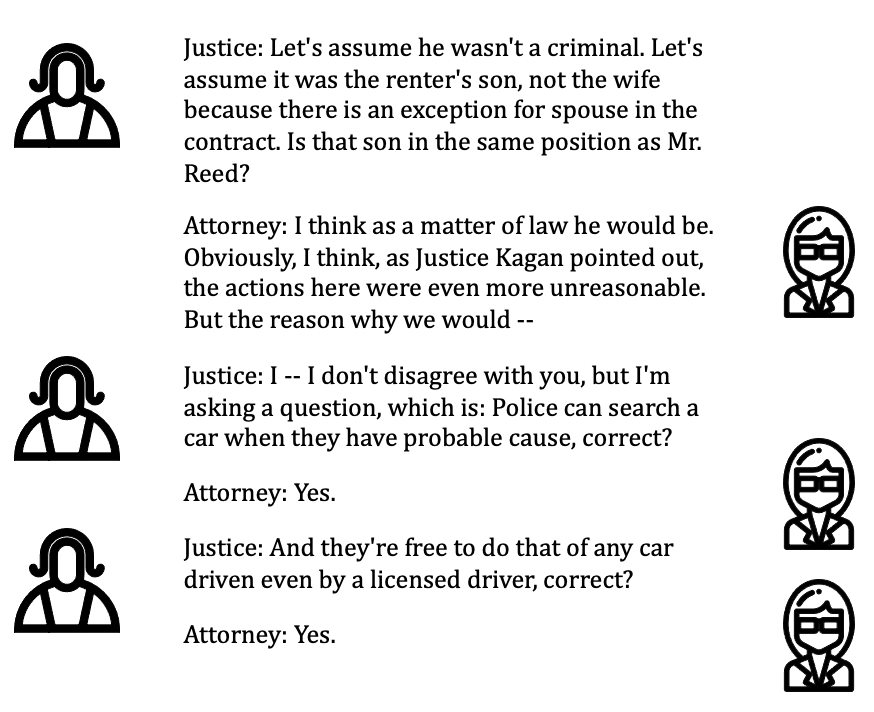}
    \caption{Justice continuously pressing attorney by making the rapid succession of her questions, cuts off attorney and then restricts him to one-word responses
before another question is initiated.}
    \label{img:perception3}
\end{figure}

\begin{figure*}[ht]
    \centering
    \subfloat[First 130 epochs]{%
        \includegraphics[width=0.48\textwidth]{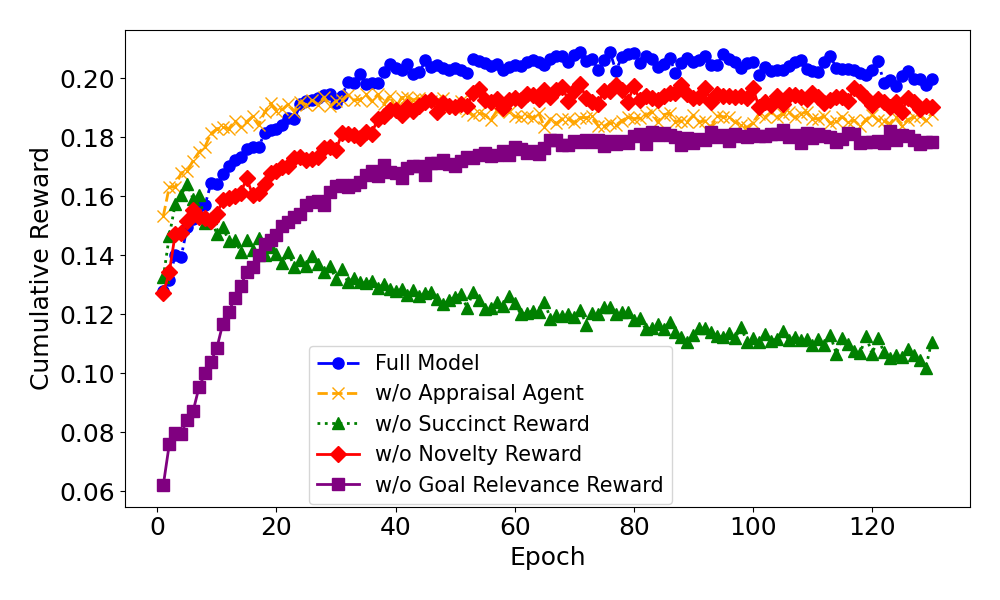}%
        \label{fig:ablation_short}%
    }
    \hfill
    \subfloat[First 1600 epochs]{%
        \includegraphics[width=0.48\textwidth]{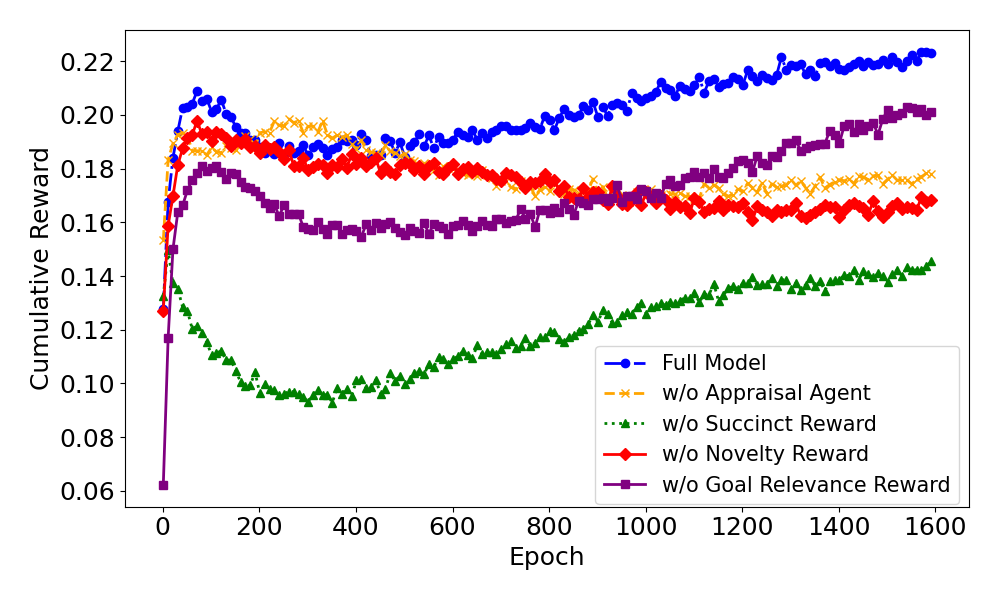}%
        \label{fig:ablation_long}%
    }
    \caption{%
    Learning Curves from Ablation Study. (a) Cumulative reward during early training stage; (b) Cumulative reward extends to longer term training (1600 epochs). The full model (blue) outperforms all ablated versions.
    }
    \label{fig:ablation}
\end{figure*}

\begin{table}[ht!]
    \centering \small
    \begin{tabular}{p{1.5cm} p{5.5cm}}
    \centering \textbf{Problem} & \centering \textbf{Snippet} \tabularnewline
    \hline
        Sub-optimal utterance & Attorney: So I -- so I don't think there's any statement in the legislative history that says we're not forcing employers to give benefits for non-work-related injuries. What -- there are three statements in the legislative history that -- that Respondent draws a negative inference from. \newline
        Justice: I'm so relieved.
\\
\hline
        Frequently \newline interception & Justice: So you really can't... there's no analytical distinction, then-- \newline Attorney: Well-- \newline Justice: --between the fact and the feeling. \newline Attorney: --That's why we believe this should be a question for the district judge, who can balance all of these factors. In your hypothetical-- \newline Justice: Yes, but even on your balancing theory I thought the judge was supposed to draw... maybe I misunderstood you. I thought the judge was supposed to draw a line between fact and feeling, and what he was supposed to be balancing-- \newline Attorney: --No, I-- \newline Justice: --was the appropriateness of admitting the fact as against other interests. \newline Attorney: --I think that's one of the things that the trial judge could be balancing, whether it's fact or feeling, but also the need for the evidence. If we had a hypothetical where the-- \newline Justice: I don't understand that, the need for the evidence?
\\ \hline
        Missing data & Attorney: That's correct. It may be applied in the discretion of the agency head... \newline Justice: (Inaudible) \newline Attorney: Yes, sir, I think there is a substantial difference and I think that's ... \\
\hline
    \end{tabular}
    \caption{Examples of Low-quality Snippets of Dataset}
    \label{low_quality}
\end{table}

\begin{table}[ht]
\footnotesize
\begin{tabular}{|p{1.8cm}|c|c|c|c|c|}
\hline
                  & CS   & PS   & OS   & PES    & Overall  \\ \hline
\makecell{\textbf{Full Model}}       & \textbf{3.99} & \textbf{4.53} & 4.32 & \textbf{4.63}  & \textbf{4.37} \\ \hline
\makecell{w/o \\Appraisal \\Agent}       & 3.83 & 3.89 & \textbf{4.42} &3.89 & 4.01         \\ \hline
\makecell{w/o Succinct \\Reward}       & 3.45 & 4.05 & 4.11 & 4.05 &  3.92         \\ \hline
\makecell{w/o Novelty \\Reward}        & 3.74 & 4.32 & 4.26 & 4.37 & 4.17          \\ \hline
\makecell{w/o Goal \\Relevance} & 3.77 &4.26&4.21 & 4.42  & 4.17         \\ \hline
\makecell{SaulLM-7B}         & 3.73 & 3.21 & 4.05 &  3  &3.5 \\ \hline
\end{tabular}
\caption{Human Evaluation}
\label{table:human_evalutaion}
\end{table}

\section{Transferability Discussion}
Here, we provide a discussion of how the framework could be adapted to other inquisitive domains.

Inquisitive conversations usually have an ultimate result. For the Supreme Court, it is a conclusion; for journalism, it is a summary; and for medical interviews, it can be highlights.

\noindent
\textbf{Appraisal taxonomy:}
To adapt to other domains, we can keep the same turn-level appraisal mechanism, but broaden the label space to a domain-general core plus domain-specific refinements. For journalism, refinements emphasize attribution and verifiability (e.g., “claim lacks source,” “timeline inconsistent,” “needs evidence/documents”). For medicine, refinements emphasize clinical completeness and safety (e.g., “missing onset/duration/severity,” “red-flag unaddressed,” “contraindication risk”). Practically, the domain-specific appraisal set can be selected from a larger universal pool, or induced with weak supervision/clustering over (question, answer, follow-up) triples—reducing reliance on handcrafted legal notions while preserving the same control interface to the dialogue policy.

\noindent
\textbf{Dialogue act hierarchy:}
The hierarchical decision structure can be broadened to a general property of inquisitive interviewing. Adaptation does not require redesigning the hierarchy; it requires swapping the act inventory. In journalism, high-level acts like clarify, verify, challenge, request evidence, reconcile contradictions, summarize leads naturally decompose into finer acts (e.g., “ask for document,” “ask for source identity,” “pin down time/place”).

Rather than hand-crafting these for each new domain, a practical adaptation path is to derive the hierarchy via hierarchical clustering of question/response embeddings or other cues, then map clusters to interpretable parent nodes while letting leaves remain domain-specific.

\noindent
\textbf{Reward design:}
The reward template also generalizes with a simple substitution: replace the Supreme Court “case conclusion” target with a domain “goal artifact” that represents the interview’s intended end product. For journalism, this can be a set of story claims/summarization of a dialogue. For medical interviews, this can be a set of highlights of it.

Generally speaking, goal-relevance then rewards answers that add content aligned with these goal elements, novelty rewards information that was not already established earlier in the conversation, and clarity is recalibrated per domain (journalism: specificity and attribution/evidence presence). Crucially, the training objective and reward combination remain unchanged—only the goal artifact and clarity proxy are swapped in this way, adaptation can be interpreted as a “plug-in” process rather than a substantial redesign.

\end{document}